\documentclass[runningheads]{llncs}
\usepackage{graphicx}
\usepackage{comment}
\usepackage{amsmath,amssymb,amsfonts}
\usepackage{adjustbox}
\usepackage{hyperref}
\usepackage[misc]{ifsym}

\begin{document}
\title{Deep Spatiotemporal Clustering: A Temporal Clustering Approach for Multi-dimensional Climate Data}

\author{Omar Faruque\inst{1}\orcidID{0009-0006-8650-4366} \and
Francis Ndikum Nji\inst{1} \and Mostafa Cham\inst{1} \and Rohan Mandar Salvi\inst{1} \and Xue Zheng\inst{2} \and  Jianwu Wang \inst{1}\orcidID{0000-0002-9933-1170}(\Letter)}
\authorrunning{O. Faruque et al.}
\tocauthor{O. Faruque et al.}
\titlerunning{Deep Spatiotemporal Clustering}
\toctitle{Deep Spatiotemporal Clustering: A Temporal Clustering Approach for Multi-dimensional Climate Data}
%
%
\institute{Department of Information Systems, University of Maryland, Baltimore County, Baltimore, MD, United States \\
\email{\{omarfaruque, fnji1, mcham2, rsalvi2, jianwu\}@umbc.edu}\\
\url{https://bdal.umbc.edu} \and
Climate Science Section, Lawrence Livermore National Laboratory, Livermore, CA, United States\\
\email{zheng7@llnl.gov}}
%
\maketitle              
\begin{abstract}
Clustering high-dimensional spatiotemporal data using an unsupervised approach is a challenging problem for many data-driven applications. Existing state-of-the-art methods for unsupervised clustering use different similarity and distance functions but focus on either spatial or temporal features of the data. Concentrating on joint deep representation learning of spatial and temporal features, we propose Deep Spatiotemporal Clustering (DSC), a novel algorithm for the temporal clustering of high-dimensional spatiotemporal data using an unsupervised deep learning method. Inspired by the U-net architecture, DSC utilizes an autoencoder integrating CNN-RNN layers to learn latent representations of the spatiotemporal data. DSC also includes a unique layer for cluster assignment on latent representations that uses the Student's t-distribution. By optimizing the clustering loss and data reconstruction loss simultaneously, the algorithm gradually improves clustering assignments and the nonlinear mapping between low-dimensional latent feature space and high-dimensional original data space. A multivariate spatiotemporal climate dataset is used to evaluate the efficacy of the proposed method. Our extensive experiments show our approach outperforms both conventional and deep learning-based unsupervised clustering algorithms. Additionally, we compared the proposed model with its various variants (CNN encoder, CNN autoencoder, CNN-RNN encoder, CNN-RNN autoencoder, etc.) to get insight into using both the CNN and RNN layers in the autoencoder, and our proposed technique outperforms these variants in terms of clustering results.  

\keywords{Temporal clustering \and Multivariate spatiotemporal climate data \and Deep neural network \and U-net.}
\end{abstract}
\section{Introduction}
Spatiotemporal data is commonly available in many disciplines such as Earth sciences, atmospheric science, and environmental science. Such data is often generated by monitoring a certain area over a period of time, which results in datasets in four dimensions (4D): time, longitude, latitude, and measured variables such as temperature and humidity. One important way to study such spatiotemporal data is to categorize the records into smaller groups by conducting unsupervised data clustering along the temporal dimension. 

There are several challenges in clustering such 4D spatiotemporal data. First, most traditional clustering algorithms like $k$-means~\cite{b1} only work on 2D tabular data and face challenges to work with 4D spatiotemporal data. Converting 4D spatiotemporal data directly into 2D tabular data will not only end up with very high dimensional data but also lose spatial and temporal patterns in the original data. Second, common dimension reduction approaches such as PCA~\cite{b7} can reduce data dimensionality before applying clustering algorithms, but such efforts \cite{b7,b8} fail to preserve the dataset's nonlinear relationship, which leads to subpar clustering accuracy. Third, recently developed deep learning-based clustering algorithms \cite{b10,b11,b12,b13} are able to learn nonlinear characteristics of the dataset. But these techniques only focused on the spatial or temporal features of the dataset, not spatiotemporal features jointly. 

To address the above challenges of high-dimensional spatiotemporal clustering, we propose a novel spatiotemporal autoencoder model drawing inspiration from the recent success of the U-net architecture~\cite{b19} in representation learning. Our model is applied to a popular climate data called ECMWF ERA5 global reanalysis product~\cite{b21} to evaluate its performance. The proposed model nonlinearly maps the input dataset with the lower dimensional hidden feature space. To make this latent feature generation more robust, we evaluated both the spatial and temporal properties of the dataset by combining CNN and LSTM layers in the encoder module. A custom clustering layer is applied to the latent features to generate clustering results. The clustering layer uses the inherent logic of the Student's t-distribution and iteratively improves the result. At the same time, the decoder module adjusts its weights to reduce the disparity between the input and reconstructed data while learning to reconstruct the high-dimensional input data from lower-dimensional latent features. From our experimental analysis, it is evident that the proposed model achieved significant improvement in cluster accuracy and holds the ability to capture the generic properties of different time series. In summary, our contributions are i) end-to-end learning of both spatial and temporal features in the same model, ii) significant improvement in clustering accuracy, and iii) iterative joint optimization of latent features and clustering assignment. Our implementation source code can be accessed at the Big Data Analytics Lab GitHub repository: \url{https://github.com/big-data-lab-umbc/multivariate-weather-data-clustering/tree/DSC}.


The remainder of the paper is structured as follows. Section~\ref{background} explains the background and definition of our clustering task. State-of-the-art related works are summarized in Section~\ref{related}. Section~\ref{method} gives a detailed description of the proposed procedure. The experimental details of the proposed model are provided in Section~\ref{experiments} along with the results and an ablation study of the proposed model. Finally, we conclude and summarize in Section~\ref{conclusion}.

\section{Background and Problem Definition}
\label{background}
\subsection{Clustering Multi-dimensional Climate Data}
Earth's climate system is a highly complicated and interconnected global system formed with a large number of dynamic components such as global temperature, ocean temperature, arctic sea ice, precipitation, wind pattern, pressure, aerosol, cloud, etc. Climate change is one of the most threatening issues because it has diverse effects on the global ecosystem and will make the weather more hazardous~\cite{b1}. Due to its high importance, many researchers focused on studying the interactions of constituent components of the climate system and the changes caused by each other. In Earth’s climate system, one of the significant components is the air-sea-cloud and their inherent interactions. The dynamic, thermodynamic, and anthropogenic processes that connect the atmosphere and oceans through clouds are referred to as air-sea-cloud interaction. Specifically, marine boundary layer clouds play an important role in air-sea-cloud interactions, as it has a strong influence to lower the sea surface and earth's temperature, and also the microphysical and dynamical characteristics of marine boundary layer clouds are sensitive to the sea surface situation~\cite{b2}. 

Observations of air-sea-cloud interactions show the tendency of high variability for a wider range of temporal and spatial scales. This is caused by different interacting atmospheric components that fuse various uncertainties in the Earth system. To untangle the effect of different components on atmospheric interactions as a function of synoptic-scale (approximately 1000 km) changes requires quantification of their interaction patterns~\cite{b3}. To achieve this goal, it is required to study different atmospheric properties of synoptic-scale regions covering a wide range of longitude and latitude over a longer period. Studying this large volume of spatial and temporal data is very complex and time-consuming for domain experts. The inherent complexity of atmospheric system study can be reduced by grouping environmental contexts based on spatial and temporal similarity, as each sub-group will demonstrate a higher small-scale perturbation and will minimize the boundary of the global atmospheric effects. 

To study the atmospheric properties over a synaptic regime we need to consider the measurements of different variables over a range of longitude and latitude. To untangle the interaction and relative effect of these components we need to quantify these atmospheric properties for a longer period. Hence, the problem size can be represented by a dataset of four dimensions: time, longitude, latitude, and variables. We can consider the task as an unsupervised spatiotemporal clustering problem of the multidimensional data, as the observations contain both the location and time-varying features. Also, the absence of the labeled dataset makes the problem an unsupervised learning task.

\subsection{Problem Definition}
The goal of the proposed model is to assign observation records of the dataset into different clusters based on the latent spatial and temporal features learned by the deep autoencoder model. Let us assume, $n$ atmospheric variables ($v_i$) are measured over a grid region covering $L$ longitudes and $W$ latitudes and stored in a vector $V=\{v_1, v_2, v_3, ..., v_n\}$. So for each time step, every grid location has $n$ values for all variables. Also, these variables are measured for $T$ different time steps, $V_i=\{v_1, v_2, v_3,..., v_n\}$, $i\in \{1, ...,T\}$. 

\vspace{0.3cm}
\textbf{Input}
\begin{center}
$Dataset = \{V_1, V_2, V_3, ..., V_T\}$
\end{center}
\begin{equation*}
\begin{split}
V_{i}&=\begin{Bmatrix}
\begin{bmatrix}
 v_{1}(1,1)&v_{1}(1,2)  & \cdots  & v_{1}(1,W) \\ 
 v_{1}(2,1)&v_{1}(2,2)  & \cdots  & v_{1}(2,W) \\ 
 \vdots &  \vdots  & \ddots  &\vdots  \\ 
 v_{1}(L,1)&v_{1}(L,2)  & \cdots  & v_{1}(L,W)
\end{bmatrix} \vspace{0.2cm}\\ 
\begin{bmatrix}
 v_{2}(1,1)&v_{2}(1,2)  & \cdots  & v_{2}(1,W) \\ 
 v_{2}(2,1)&v_{2}(2,2)  & \cdots  & v_{2}(2,W) \\ 
 \vdots &  \vdots  & \ddots  &\vdots  \\ 
 v_{2}(L,1)&v_{2}(L,2)  & \cdots  & v_{2}(L,W)
\end{bmatrix} \\
 \vdots \\
\begin{bmatrix}
 v_{n}(1,1)&v_{n}(1,2)  & \cdots  & v_{n}(1,W) \\ 
 v_{n}(2,1)&v_{n}(2,2)  & \cdots  & v_{n}(2,W) \\ 
 \vdots &  \vdots  & \ddots  &\vdots  \\ 
 v_{n}(L,1)&v_{n}(L,2)  & \cdots  & v_{n}(L,W)
\end{bmatrix} 
\end{Bmatrix}
\end{split}
\end{equation*}
Here, $V_i$ represents one observation, $v$ means one variable of an observation, $i \in \{1, ..., T\}$, $T$ is the number of time steps, $n$ is the number of atmospheric variables, $L$ is the longitude and $W$ is the latitude. 

\vspace{0.3cm}
\textbf{Output}

In particular, the proposed model will categorize the dataset $\{V_1, V_2, V_3, ...,$ $V_T\}$ into $k$ clusters: $C_1$, $C_2$, $C_3$, ..., $C_k$, where $k<T$, so that the members of a cluster are more similar to each other and dissimilar from the members of all other clusters. Formally: 
\begin{equation*}
    C_1=\{V_{C_1}^1, V_{C_1}^2, ..., V_{C_1}^{n_1}\}, C_2=\{V_{C_2}^1, V_{C_2}^2, ..., V_{C_2}^{n_2}\}, ..., C_k=\{V_{C_k}^1, V_{C_k}^2, ..., V_{C_k}^{n_k}\} 
\end{equation*}
\begin{equation*}
V_{C_j}^{i} \in V, i\in \{1, ..., n_j\}, j\in \{1, ..., k\} 
\end{equation*}
$n_j$= number of observations of cluster $j$. 
\begin{equation*}
\bigcup_{j=1}^{k}C_j=V \textrm{ and } C_j\cap C_l=\varnothing
\end{equation*}
Here, $j\neq l$ and $(j, l)\in \{1, ..., k\}$.

\section{Related Works}
\label{related}

\textbf{Traditional Clustering Algorithms}. Unsupervised clustering is an extensively explored branch of machine learning. A large number of clustering algorithms have been developed by focusing on feature selection, similarity measure, grouping process, and cluster validation strategies. The $k$-means algorithm~\cite{b14} is one of the most popular clustering methods. This method is very effective for a large number of unsupervised clustering problems and computationally very efficient. However, the $k$-means algorithm performs better for low dimensional data than for high dimensional data. Density-based spatial clustering of applications with noise (DBSCAN)~\cite{b15} is another popular unsupervised clustering technique that works mainly focusing on the number of neighboring data points. This method can work with any shape of the cluster and it automatically identifies a suitable number of clusters for the dataset. But the DBSCAN performs poorly if the dataset's density varies by a large margin. The agglomerative hierarchical clustering technique~\cite{b16} is a bottom-up approach, which initially considers each data point as a separate cluster and iteratively reduces the number of clusters by aggregating two similar clusters into a new cluster until the desired number of clusters is found. The time and space complexity of this method is high and also does not perform well for high-dimensional datasets. To apply the $k$-means, DBSCAN, and hierarchical clustering algorithms on a high dimensional dataset, several variations were proposed in \cite{b17,b18} using dimensionality reduction. But these dimensionality reduction methods create a linear mapping between the input data space to the low-dimensional embeddings and fail to address nonlinear complex data. 

\textbf{Deep Learning based Clustering}. In recent years, deep learning models have been applied dominantly to learn the nonlinear and complex patterns of the input data \cite{b4,b5,b6}. Deep Embedded Clustering (DEC)~\cite{b10} is an unsupervised clustering method that maps the input data into a low-dimensional embedding space using a deep neural network. Starting with an initial representation and cluster assignment DEC iteratively optimizes both by using Kullback-Leibler (KL) divergence loss. Although DEC generates better classification results but loses the spatial details of the image. To solve this problem the Clustering-Augmented Segmentation (CAS)~\cite{b12} model proposed an autoencoder-based representation learning method using U-net architecture. CAS model generates the land segmentation using clustering loss and data reconstruction loss. Deep Temporal Clustering Representation (DTCR)~\cite{b13} is a novel unsupervised clustering model for time series data that used the bidirectional recurrent neural network to learn temporal representation. In this model data reconstruction and $k$-means loss are combined into the seq2seq model to generate cluster-specific representations. To cluster the time series data, the Deep Temporal Clustering (DTC)~\cite{b11} model utilizes an autoencoder to reduce the input data in a lower dimensional space and then optimizes the clustering objective. The DTC model used temporal reconstruction loss and clustering loss to optimize the model parameters and clustering results. The authors in~\cite{b26} also used autoencoder to generate latent representations from sentence embeddings of short text and then performed clustering using latent features. One limitation of these models is that they either concentrate on temporal features or pixel-level similarities. So these approaches lack a general methodology for learning effective latent representation and unsupervised clustering of spatiotemporal data. 

\section{Proposed Methodology}
\label{method}
\begin{figure}
\includegraphics[width=\textwidth]{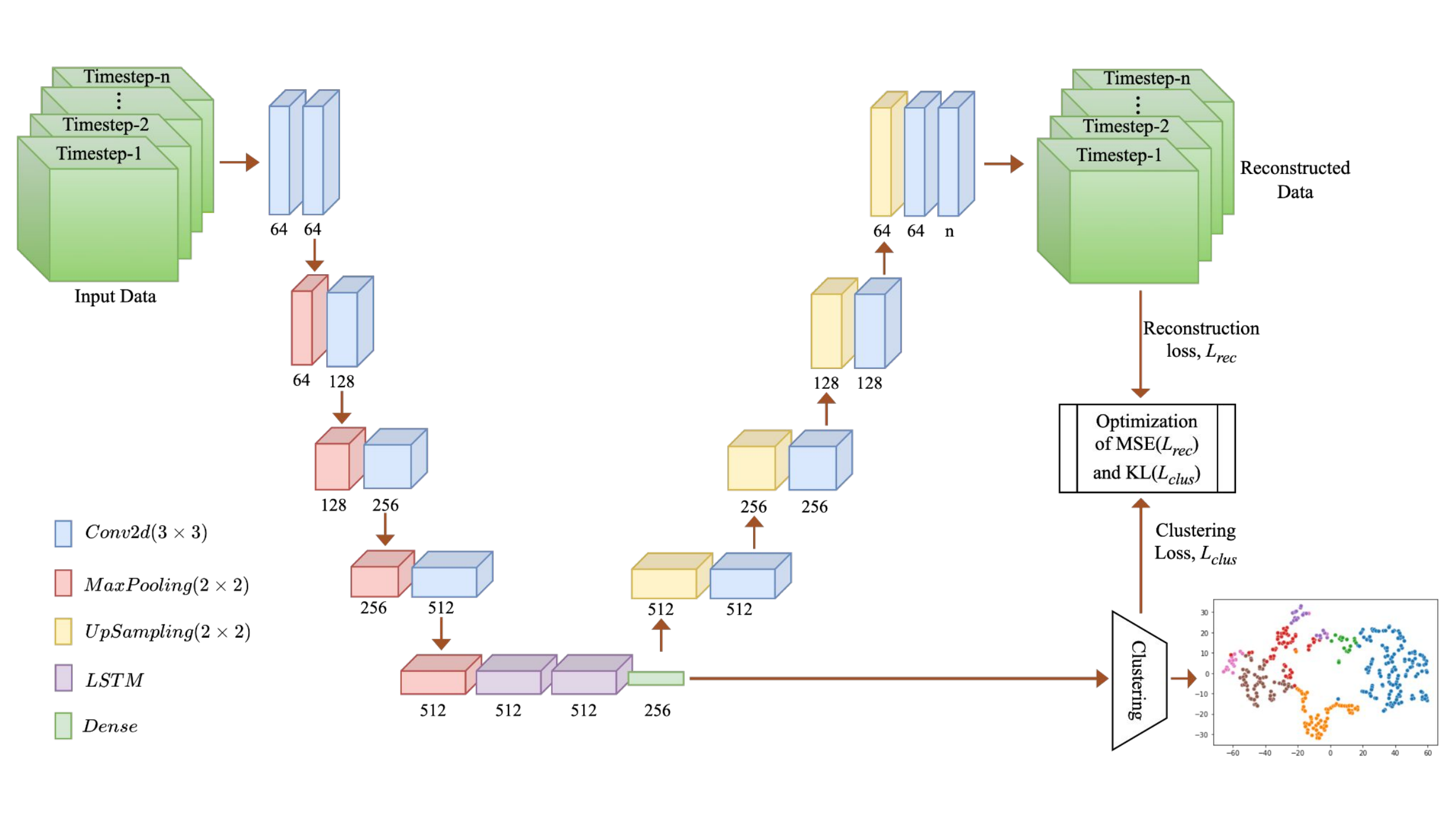}
\caption{Illustration of the proposed deep spatiotemporal clustering model architecture (best viewed in color). The number at the bottom of each block means the number of feature layers.} 
\label{model}
\end{figure}

In this paper, we propose a novel autoencoder model for clustering unlabeled spatiotemporal datasets. To capture the hidden spatial and sequential features from the dataset, we integrate the Convolution Neural Network (CNN) and Long Short Term Memory (LSTM) in the model. The proposed model is illustrated in Figure~\ref{model}. The encoder module transforms the input data into a latent representation ignoring any cluster-specific features. Then, using that latent representation, the decoder reconstructs the input data. The clustering objective is also incorporated into the model to train it for better outcomes in representation learning and clustering.

\subsection{Overview of our Deep Spatiotemporal Clustering (DSC) Approach}

Given the spatiotemporal dataset $V$ such that $V_t \in \mathbb{R}^{lon \times lat \times n}$, where $n$ is the number of the time series variables and  $t \in \{1, ..., T\}$. To sub-group $T$ timesteps into $k<T$ clusters the encoder module transforms the input data $V$ into a high-level latent representation $E$ with reduced data dimension. The mapping from the input data space to the latent feature space is a nonlinear function $f_{en}:= V \rightarrow E$. Here $E \in \mathbb{R}^{m}$ is an $m$-dimensional high-level representation of all the variables at each timestep.

\begin{equation}
E_t = f_{en}(V_t), t = \{1, ..., T\}
\end{equation}
For unsupervised clustering of the high dimensional data, it is very crucial to learn a latent representation that well represents the input data. To achieve this requirement in the nonlinear encoder, we used 2D convolution, which extracts the hidden spatial features from the dataset. To extract latent features at different scales and reduce the dimensionality, we used several stacks of 2D convolution and max pooling layers. For further processing, dimensionality reduction is essential to prevent extremely long sequences, which might result in weak performance, and to overcome the curse of dimensionality. The activations of the convolution layers are then applied to the LSTM layers to learn temporal changes. LSTM layers cast the latent representation into a more compact space in the temporal direction. Finally, using the latent features of the encoder model, the clustering layer assigns the set of data points $V_i$, $i=\{1, ..., T\}$, into $k$ clusters with distinct spatiotemporal features. The decoding process of the proposed model performs the inverse operation of the encoder in a nonlinear fashion, $f_{de}:= E \rightarrow \widehat{V}$. It maps the compact latent features of the encoder function into a new feature space that is identical to the input dataset, $\widehat{V} \approx V$. The final reconstruction of the decoder module $\widehat{V} \in \mathbb{R}^{lon \times lat \times n}$ is defined as
\begin{equation}
\widehat{V}_t = f_{dc}(E_t), t = \{1, ..., T\}
\end{equation}
By reconstructing the input data from latent representations the decoder verifies that the behaviors of the spatiotemporal sequences are well preserved after dimensionality reduction. In the proposed model the decoder process is constructed using stacked upsampling and convolution methods. This helps the decoder network to learn effective reconstruction parameters and to reduce the difference between the reconstructed and input data.    

The proposed model clusters data points by simultaneously learning a set of clusters in the latent feature space by minimizing two objective functions jointly (more at Section 4.3). The first objective function emphasizes generating well-separated groups in the latent space by reducing the clustering loss. The second objective function tries to reduce the mean square error of the reconstructed data from the input dataset, called reconstruction loss. The collective optimization of both objective functions guides the autoencoder to extract efficient spatiotemporal features that are best suited to distribute the input data into $k$ categories. The property of learning spatial and temporal features through end-to-end optimization establishes our autoencoder model prominent from the state-of-the-art unsupervised clustering algorithms. The traditional models either emphasize the clustering loss or the reconstruction. Some models optimize both the clustering and reconstruction loss but loosely concentrate on the temporal and spatial features together. The experiments in Section~\ref{experiments} will show the improvement of our proposed model over these related unsupervised clustering algorithms.

\subsection{Clustering Assignment}
At the beginning of the model training process, the input data $V_i$ is applied to the freshly initialized autoencoder to obtain the first generation latent representation $E_i$ which is a vector of 256 dimensions. To estimate the initial cluster centroids $C_j$, $j=\{1, ..., k\}$ the $k$-means algorithm applied on these latent features $E_i$. Then the proposed unsupervised clustering layer iteratively improves the clustering by a two-step procedure inspired by the previous work~\cite{b10}. In the first step, a soft assignment of each data point $V_i$ to all clusters is computed based on the similarity between the cluster centroid $C_j$ and latent representation $E_i$. In the second step, the nonlinear mapping $E_i$ and the cluster centroids $C_j$ are refined by using the clustering loss function, which learns from high-confidence cluster assignment of the current target distribution. Each data point's soft assignment is computed using the Student’s t-distribution~\cite{b9}, which measures the similarity between the embedded representation $E_i$ and the cluster centroids $C_j$. The soft assignment of data point $i$ to cluster $j$ is given by:
\begin{equation}
q_{ij}=\frac{(1+||E_{i}-C_{j}||^{2}/\alpha )^{-\frac{\alpha +1}{2}}}{\sum_{l=1}^{k}(1+||E_{i}-C_{l}||^{2}/\alpha )^{-\frac{\alpha +1}{2}}}
\end{equation}
Here $q_{ij}$ represents the probability of assigning the $i$'th data point to the $j$'th cluster and $\alpha$ is the degree of the freedom of the Student’s t-distribution which is set as $1$ in our proposed model. At each iteration, the value of $q_{ij}$ is computed based on the previous iteration's cluster centroids. As $q_{ij}$ are soft assignments, the predictions are strengthened using a target distribution of data points. The target distribution puts more attention on data points with the high-confidence cluster assignment and normalizes the loss contribution of each centroid to prevent the distortion of the latent feature space. The target assignment $p_{ij}$ is given by:
\begin{equation}
p_{ij}=\frac{{q_{ij}^{2}}/{\sum_{i=1}^{n}q_{ij}}}{\sum_{l=1}^{k}({q_{il}^{2}}/{\sum_{i=1}^{n}q_{il}})}
\end{equation}
Once the values of $q_{ij}$ and $p_{ij}$ are computed for all data points, the cluster centroids $C_j$ are refined using gradient descent. Each data point's cluster assignment gets recalculated following the change of the cluster centroids, and it is then compared to the previous cluster assignment. If there is no change in the cluster assignment of any data point for several consecutive iterations, the model terminates the optimization process and returns the cluster assignment from that iteration.    

\subsection{Joint Optimization}
In the proposed unsupervised clustering process, we used the autoencoder model to reconstruct the input data $V$ and the clustering layer for cluster assignment from latent features $E_i$ of the input data $V_i$. During the learning process, we jointly optimized the autoencoder weights and the clustering by minimizing the mean squared error and the Kullback-Leibler (KL) divergence loss respectively. This optimization task is implemented using the Stochastic Gradient Descent (SGD) method with momentum. SGD guides the autoencoder model to learn efficient latent embeddings to capture distinctive and representative features of the input dataset. The reconstruction loss of the model is computed by the mean squared error, which is given by
\begin{equation}
L_{rec}=min(\frac{1}{T}\sum_{i=1}^{T}\left \| V_{i}-\widehat{V_{i}} \right \|_{2}^{2})
 \end{equation}
Here  $V_i \in \mathbb{R}^{lon \times lat \times n}$ is the input data and  $\widehat{V_i} \in \mathbb{R}^{lon \times lat \times n}$ is the output of the autoencoder model. The cluster centroids and eventually the cluster assignment is iteratively refined in the model by minimizing the clustering loss. The KL divergence is used to compute the clustering loss between the soft assignment $q_{ij}$ and the target distribution $p_{ij}$ of each data point to clusters. The formula of the KL divergence loss is
\begin{equation}
\begin{split}
L_{clus}=min(KL(P\parallel Q))
=min(\frac{1}{T}\sum_{i=1}^{T}\sum_{j=1}^{k}p_{ij}log\frac{p_{ij}}{q_{ij}})    
\end{split}
\end{equation}
Here $T$ is the number of timestamp/data points and $k$ is the number of clusters, $k<T$. Finally, the combined loss for training the proposed unsupervised clustering models is defined by:
  
\begin{equation}
\begin{split}
   L&=L_{clus}+L_{rec}\\
   & =\frac{1}{T}\sum_{i=1}^{T}\left (\left \| V_{i}-\widehat{V_{i}} \right \|_{2}^{2} +\sum_{j=1}^{k}p_{ij}log\frac{p_{ij}}{q_{ij}} \right)
\end{split}
\end{equation}

\section{Experiments}
\label{experiments}
Our proposed model was implemented using the Python deep-learning libraries Keras 2.11 and TensorFlow 2. All the baseline models and proposed models were tested in the Google Colab notebook with 12 GB GPU memory support. For a fair comparison, we used the same Keras library for all baseline models. We executed each model 20 times with random initialization and reported the best evaluation result. 



\subsection{Dataset and Data Preprocessing}
For this study, we use the open-access atmospheric reanalysis data  from European Centre for Medium-Range Weather Forecasts (ECMWF) ERA-5 global reanalysis product~\cite{b21}. The ERA5 atmospheric dataset contains 31 km high-resolution reanalysis and a reduced resolution ten-member ensemble~\cite{b22}. The ensemble is required for the data assimilation procedure. Seven atmospheric reanalysis variables from the ERA5 dataset were selected for this study. These are Sea Surface Temperature (unit: $k$, and range: 285 to 300), Surface Pressure (unit: $Pa$, and range: 98260 to 103788), Surface Sensible Heat Flux (unit: $J/m^2$, and range: -674528 to 200024), Surface Latent Heat Flux (unit: $J/m^2$, and range: -1840906 to 90131), 2-meter Air Temperature (unit: $k$, and range: 281 to 299), 10-meter U-component of Wind (unit: $m/s$, and range: -16 to 19), and 10-meter V-component of Wind (unit: $m/s$, and range: -15 to 16). These variables are included in the dataset based on their impact on the air-sea-cloud interaction system and were measured in a latitude-longitude grid of (41x41). Temporally, the dataset covers one year period and one observation per day. If we directly convert the data into a 2D tabular data frame, the total feature count for each record would be 11,767 (41x41x7), which is clearly a high-dimensional dataset. 

After exploring the dataset we found the presence of null values. These null values may occur due to the sensor malfunction or any physical conditions. The null values are replaced by the overall mean of the dataset because the replacement with any neighboring values may create a different pattern in the data, which obviously will change the actual behavior of the variable. From the description of the dataset, it is clear that the value ranges of the variables are different. It is also necessary to make the features of the dataset into the same scale for better feature learning through a deep neural network~\cite{b23}. For this dataset, we utilized the standard Min–Max Normalization (MMN) normalization to rescale all features within the range of 0 to 1. Each variable $V_i$ rescaled using the following formula:
\begin{equation}
    V_i=\frac{V_i-min(V_i)}{max(V_i)-min(V_i)}
\end{equation}

\subsection{Baseline Methods}
We compared the result of the proposed model with the $k$-means algorithm and the hierarchical clustering algorithm. These methods are the most popular methods for unsupervised clustering. The original 4D input dataset is transformed into a 2D matrix with one row per timestep to apply these algorithms. To contrast with our suggested approach, we additionally used the DEC~\cite{b10}, DTC~\cite{b11}, and DTCR~\cite{b13} algorithms from the deep-learning family and these algorithms employed a deep neural network for dimensionality reduction after transforming each observation's data from a multi-dimension matrix to a one-dimensional row. The low-dimensional latent features produced by the deep neural network were then subjected to the clustering algorithm. The comparison between baseline models and the proposed model is depicted in Table~\ref{tab3}.  

\begin{table}
\caption{Comparison among baseline methods.}
\label{tab3}
\begin{adjustbox}{width=\textwidth}
\begin{tabular}{|c|c|c|c|c|}
\hline
 \textbf{Method} & \textbf{Input Shape} & 
\textbf{Technology} & \textbf{Pretraining} \\
\hline
\textit{k}-means & 2D Matrix & Traditional distance/similarity-based model & Not required \\
\hline
Hierarchical Clustering & 2D Matrix & Traditional distance/similarity-based model & Not required \\
\hline
DEC & 2D Matrix & Deep learning model with dense layers & Required \\
\hline
DTC & 2D Matrix & Deep learning model with LSTM layers & Required \\
\hline
DTCR  & 2D Matrix &	Deep learning model with GRU layers & Required  \\
\hline
DSC (ours)  & 4D Matrix &	Deep learning model with CNN and LSTM layers & Not required  \\
\hline
\end{tabular}
\end{adjustbox}
\end{table}

\subsection{Evaluation Metrics}
We used Average Intercluster Distance, Average Variance, ${RMSE}_{mean}$, Silhouette coefficient, and Davies-Bouldin score evaluation metric to measure the performance of our experiments. As the dataset does not have any ground truth cluster values, intercluster distance will give a reasonable estimation of the quality of the clustering by representing the proximity of generated clusters. Intercluster distance is measured as the minimum distance between any two data points belonging to two different clusters. 
\begin{equation}
    \Delta _{inter} (C_a,C_b)={min}_{C_a\neq C_b, X\in C_a, Y\in C_b} d(X, Y)
\end{equation}
Here $C_a$ and $C_b$ are two different clusters, $X \in C_a$ means every observation of cluster $C_a$, and similarly $Y \in C_b$ means every observation of cluster $C_b$. 

By measuring the variance of the clusters it is possible to evaluate the homogeneity and compactness of the clusters. Also, the distribution of time series over clusters should be carried out to minimize intracluster variance. The variance of each cluster will be computed as an evaluation criterion of the proposed method using the following equation  
\begin{equation}
    {Var}_{C_j}= \frac{1}{n_{C_j}} \sum_{X\in C_j}^{} \left \| X- {\mu}_{C_j} \right \|^{2}
\end{equation}
Here ${n_{C_j}}$ is the total number of observations in the cluster $C_i$ and ${\mu}_{C_j}$ is mean of the cluster. 

The root-mean-squared error (RMSE) is another common measure of clustering quality evaluation. For each observation, the error is the distance between every observation and its associated cluster centroid. The total clustering error is therefore the sum of the errors associated with each data point. Then we take root over the average of the total error. 
\begin{equation}
    {RMSE}_{mean}= \sqrt{
    \frac{1}{n} \sum_{j=1}^{k} \sum_{X \in C_j}^{} \left \| X- {\mu}_{C_j} \right \|^{2}
    } 
\end{equation}
Here $n$ is the length of the dataset, $k$ is the number of clusters, $X$ is the member of cluster $C_j$, and ${\mu}_{C_j}$ is the mean of the cluster $C_j$. The lower ${RMSE}_{mean}$ score means the members of each cluster are very similar and close to the cluster centroid. On the other hand, the higher ${RMSE}_{mean}$ means the opposite and is not desired from a clustering algorithm. 

Besides the above metrics, we also used the Silhouette coefficient and Davies-Bouldin score as additional metrics. The silhouette coefficient~\cite{b24} is applied to measure the cohesion and separation of the generated clusters. This is defined in the range $[-1, 1]$, where positive numbers denote a significant degree of cluster separation. Moreover, negative values show that the clusters are jumbled together (i.e., an indication of overlapping clusters). The data are said to be evenly dispersed when the silhouette coefficient is zero. 
The Davies-Bouldin score~\cite{b25} measures the similarity of various formed clusters and a lower value indicates better clustering, with zero being the lowest attainable value. 

\subsection{Experiment Results}
The comparison of the proposed method with baseline algorithms is presented in Table~\ref{tab1}. The best results for each metric are in bold. We initialized the cluster number hyperparameter as 7 for clustering the dataset. To compute these evaluation results we trained each model 20 times and picked the best result from them. The results of DEC and DTC methods are generated by running their published codes. The hyperparameters of these models like batch size, learning rate, maximum iteration, etc. are finetuned to get better results on the applied dataset. The $k$-means and hierarchical clustering method are taken from sklearn python library and applied to preprocessed data without any dimension reduction. As shown in Table~\ref{tab1}, the proposed DSC algorithm outperforms all the baseline methods except that the DEC algorithm~\cite{b10} has a lower ${RMSE}_{mean}$ score. Note that the DEC method does not consider spatial features and uses a greedy pretraining process. In contrast, our proposed method achieved ${RMSE}_{mean}$ close to DEC without any preprocessing step and significant improvement for the other four evaluation metrics.             

\begin{figure}
\includegraphics[width=\textwidth]{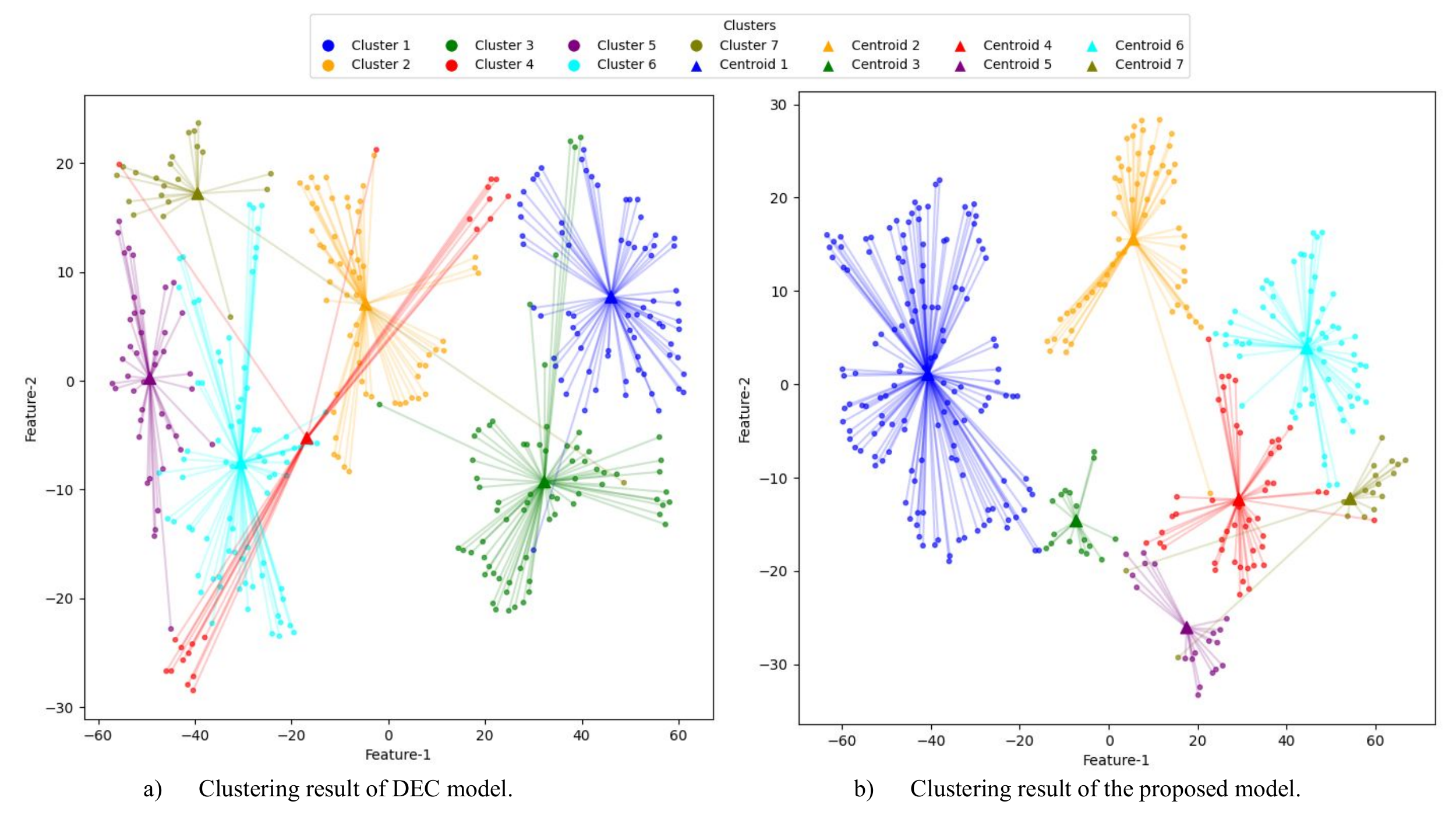}
\caption{Visualization of the clusters formed by DEC and our proposed model (DSC). The generated latent features are subjected to the t-SNE method to decrease dimension for visualization. The colors indicate the cluster label and centroid of each cluster pointed by the triangle.} 
\label{visualization}
\end{figure}

\begin{table}
\caption{Evaluation result comparison of the proposed DSC method with baseline methods.}
\label{tab1}
\begin{adjustbox}{width=\textwidth}
\begin{tabular}{|c|c|c|c|c|c|}
\hline
\multicolumn{1}{|p{3cm}|}{\centering \hspace{0.1cm}\\ \textbf{Method}} & \multicolumn{1}{|p{2cm}|}{\centering \textbf{Silhouette}\\ \textbf{Coefficient}}& 
\multicolumn{1}{|p{2cm}|}{\centering \textbf{Davies-Bouldin} \\ \textbf{Score} }
& 
\multicolumn{1}{|p{2cm}|}{\centering \hspace{0.1cm}\\ \textbf{RMSE\textsubscript{mean}} } &
\multicolumn{1}{|p{2cm}|}{\centering \textbf{Avg Inter}\\  \textbf{Cluster Distance}} &  \multicolumn{1}{|p{2cm}|}{\centering \textbf{Average}\\ \textbf{Variance}} \\
\hline
\textit{k}-means & 0.25 & 1.6867 & 13.66 & 7.00 & 0.0451 \\
\hline
Hierarchical Clustering & 0.23 & 1.5669 & 13.99 & 7.78 & 0.0453 \\
\hline
DEC & 0.26 & 1.6080 &	\textbf{13.62} & 6.75	& 0.0448 \\
\hline
DTC & 0.28 &   1.6850 &	14.70 &	6.52 &	0.0457 \\
\hline
DTCR  & -0.10 &	7.2765 &	21.14 &	4.79 &	0.0504 \\
\hline
DSC (ours) & \textbf{0.37} &	\textbf{1.4348} &	13.79 &	 \textbf{8.16} &	\textbf{0.0430} \\
\hline
\end{tabular}
\end{adjustbox}
\end{table}
In Figure~\ref{visualization}, we visualize the clusters generated by the DEC and the proposed DSC method. As the latent feature space is not suitable for visualization we use t-SNE~\cite{b9} to reduce the dimension of the feature space. The cluster plot of the DEC method shows some clear overlap between the three clusters represented by red, orange, and cyan, even though the DEC method achieved a better ${RMSE}_{mean}$ score. On the other hand, in the plot of our proposed model (DSC), all clusters are reasonably separated from each other except for some data points located far distant from the original cluster centroids. There may be one possible argument that due to the application of t-SNE dimensionality reduction, the clusters are overlapped in the DEC plot. To scrutinize that possibility, the visualization is generated multiple times, and the resultant cluster plotting reveals similar overlapping each time.                  

\subsection{Ablation Study}
To verify the impact of the spatiotemporal autoencoder of the proposed model on clustering, a comparative study of this model with its different variants is shown here. The quantitative results of these models are illustrated in Table~\ref{tab2}. In the CNN models, we only utilized convolutional neural network layers available in the Keras library and the same custom clustering layer as the proposed model. On the other hand, the CNN-LSTM Encoder model only uses the encoder part of the proposed DSC model while other functionality and parameters remain the same in both models. To gain the best performance from these models we tuned some hyperparameters like batch size, activation function, learning rate, and initial weight initialization so that we can get the actual impact of the autoencoder and custom clustering layer. The ablation study results show that the integration of LSTM layers with the CNN layers in the encoder module improves the average cluster distance and Davies-Boulding score for a significant magnitude, but the CNN encoder model achieves a little lower ${RMSE}_{mean}$. Out of these four variations, the proposed CNN-LSTM autoencoder model generated a better Silhouette Coefficient, Davies-Bouldin Score, average cluster distance, and variance. In summary, the proposed model increases the similarity of the observations in the formed clusters with better cluster separation.      

\begin{table}
\caption{Ablation results of the proposed DSC method.}
\label{tab2}
\begin{adjustbox}{width=\textwidth}
\begin{tabular}{|c|c|c|c|c|c|}
\hline
\multicolumn{1}{|p{4.5cm}|}{\centering \hspace{0.1cm}\\ \textbf{Method}} & \multicolumn{1}{|p{2cm}|}{\centering \textbf{Silhouette}\\ \textbf{Coefficient}} & 
\multicolumn{1}{|p{2cm}|}{\centering \textbf{Davies-Bouldin} \\ \textbf{Score} } & 
\multicolumn{1}{|p{2cm}|}{\centering \hspace{0.1cm}\\ \textbf{RMSE\textsubscript{mean}}} &
\multicolumn{1}{|p{2cm}|}{\centering \textbf{Avg Inter}\\  \textbf{Cluster Distance}} &  \multicolumn{1}{|p{2cm}|}{\centering \textbf{Average}\\ \textbf{Variance}} \\
\hline
CNN Encoder & 0.35 & 1.4769 &	\textbf{13.74} &	7.31 &	0.0449 \\
\hline
CNN Autoencoder & 0.31 &	1.6404 &	13.96 &	7.58 &	0.0458 \\
\hline
CNN-LSTM Encoder & 0.36 &	1.4382 &	13.82 &	8.10 &	0.0450 \\
\hline
CNN-LSTM Autoencoder (DSC) & \textbf{0.37} &	\textbf{1.4348} &	13.79	& \textbf{8.16} &	\textbf{0.0430} \\
\hline
\end{tabular}
\end{adjustbox}
\end{table}

\section{Conclusions}
\label{conclusion}
In this paper, we proposed a novel deep learning-based model, called Deep Spatiotemporal Clustering (DSC) model, to cluster high-dimensional spatiotemporal data without any supervision from previous knowledge. This model is able to learn a fine-level latent representation of the data using both spatial and temporal features, and extracting better cluster structure from the complex dataset. The joint optimization of the clustering and representation objectives using the custom clustering layer helps the model to gain better performance. The comparison with baseline unsupervised models clearly illustrates the superior clustering quality and effectiveness of the DSC model. Also, the impact of the proposed model can be perceived from the quantitative analysis of the ablation study. 

For future work, we plan to apply this model to additional high-dimensional datasets including ones from other domains. We also plan to study whether incorporating domain knowledge into the end-to-end training process could improve the results further and/or make the proposed model more robust to the domain-specific application.

\section*{Acknowledgment}

This work was supported by the DOE Office of Science Early Career Research Program. This work was performed under the auspices of the U.S. Department of Energy (DOE) by LLNL under contract DE-AC52-07NA27344. LLNL-CONF-846980. Faruque and Wang were also partially supported by grant OAC-1942714 from the U.S. National Science Foundation (NSF) and grant 80NSSC21M0027 from the U.S. National Aeronautics and Space Administration (NASA).

\bibliographystyle{splncs04}
\bibliography{references}
\end{document}